\documentclass[journal]{IEEEtran}

\ifCLASSINFOpdf
\else
\fi
\usepackage{amsmath}
\usepackage{graphicx}
\usepackage{amsmath}
\usepackage{times}

\usepackage{soul}
\usepackage{url}
\usepackage[hidelinks]{hyperref}
\usepackage[utf8]{inputenc}
\usepackage[small]{caption}
\usepackage{graphicx}
\usepackage{amsmath}
\usepackage{booktabs}
\usepackage{amssymb}
\usepackage{algorithmic} 
\usepackage{xcolor} 
\usepackage{tabularx}
\usepackage{microtype}
\usepackage{graphicx}
\usepackage{booktabs} 
\usepackage{amssymb}
\usepackage{graphicx}
\usepackage{subfigure}

\newtheorem{corollary}{Corollary}
\newtheorem{lemma}{Lemma}
\newtheorem{definition}{Definition}

\usepackage[ruled,linesnumbered]{algorithm2e}
\usepackage{float}
\usepackage{hyperref}
\usepackage[OT1]{fontenc}

\begin{document}
%
\title{Multi-Agent Feedback Enabled Neural Networks for Intelligent Communications}

%
%
%

\author{\IEEEauthorblockN{
Fanglei Sun, 
Yang Li, ~\textit{Graduate Student Member, IEEE}~,
Ying Wen,
Jingchen Hu,\\
Jun Wang,
Yang Yang,~\textit{Fellow, IEEE}~, 
Kai Li
}

\thanks{This paper is accepted by  IEEE TRANSACTIONS ON WIRELESS COMMUNICATIONS. This work was supported in part by the National Key Research and Development Program of China under Grant 2020YFB2104300, in part by the Major Key Project of Peng Cheng Laboratory (PCL) under Grant PCL2021A15, and in part by the Joint Funds of the National Natural Science Foundation of China under Grant U21B2002. (Corresponding author: Yang Yang.)}
\thanks{
Fanglei Sun is with the School of Creativity and Art, ShanghaiTech University, Shanghai 201210, China (e-mail: flsun@shanghaitech.edu.cn).}
\thanks{
Yang Li is with the School of Information Science and Technology and the School of Creativity and Art, ShanghaiTech University, Shanghai 201210,
China (e-mail: liyang2@shanghaitech.edu.cn). }
\thanks{
Ying Wen is with the John Hopcroft Center for Computer Science, Shanghai Jiao Tong University, Shanghai 200240, China (e-mail: ying.wen@sjtu.edu.cn).}
\thanks{
Jingchen Hu is with the the Institute of Mathematical Sciences,ShanghaiTech University, Shanghai 201210, China (e-mail:
hujch@shanghaitech.edu.cn).}
\thanks{
Jun Wang is with the Department of Computer Science, University College London, London WC1E 6BT, U.K. (e-mail: jun.wang@cs.ucl.ac.uk).}
\thanks{
Yang Yang is with ShanghaiTech University, Shanghai 201210, China, also with Terminus Group, Beijing 100027, China, also with the Peng Cheng Laboratory, Shenzhen 518055, China, and also with Shenzhen SmartCity Technology Development Group Company Ltd., Shenzhen 518046, China (e-mail: dr.yangyang@terminusgroup.com).}
\thanks{
Kai Li is with the School of Information Science and Technology, ShanghaiTech University, Shanghai 201210, China (e-mail: likai@shanghaitech.edu.cn).
}
}

\maketitle

\begin{abstract}

In the intelligent communication field, deep learning (DL) has attracted much attention due to its strong fitting ability and data-driven learning capability. Compared with the typical DL feedforward network structures, an enhancement structure with direct data feedback have been studied and proved to have better performance than the feedfoward networks. However, due to the above simple feedback methods lack sufficient analysis and learning ability on the feedback data, it is inadequate to deal with more complicated nonlinear systems and therefore the performance is limited for further improvement. In this paper, a novel multi-agent feedback enabled neural network (MAFENN) framework is proposed, which make the framework have stronger feedback learning capabilities and more intelligence on feature abstraction, denoising or generation, etc. Furthermore, the MAFENN frame work is theoretically formulated into a three-player Feedback Stackelberg game, and the game is proved to converge to the Feedback Stackelberg equilibrium. The design of MAFENN framework and algorithm are dedicated to enhance the learning capability of the feedfoward DL networks or their variations with the simple data feedback. To verify the MAFENN framework's feasibility in wireless communications, a multi-agent MAFENN based equalizer (MAFENN-E) is developed for wireless fading channels with inter-symbol interference (ISI). Experimental results show that when the quadrature phase-shift keying (QPSK) modulation scheme is adopted, the SER performance of our proposed method outperforms that of the traditional equalizers by about 2 dB  in linear channels. When in nonlinear channels, the SER performance of our proposed method outperforms that of either traditional or DL based equalizers more significantly, which shows the effectiveness and  robustness of our proposal in the complex channel environment. 
  
\end{abstract}


%
\IEEEpeerreviewmaketitle

\section{Introduction}
\label{introduction}
\
%
%
%
%

Intelligent communication is considered to be one of the key directions of wireless communication development after 5G. Due to its powerful nonlinear mapping and distribution processing capability, DL is being considered as a very promising tool to attack the big challenge in wireless communications and networks. It can be used to provide better channel modeling and estimation in millimeter and terahertz bands; to select a more adaptive modulation in massive multiple-input and multiple-output (MIMO) technology; and to offer a more practical solution for intelligent network optimization. DL was proposed by Hinton et al. in 2006 and is a branch of machine learning (ML). It has already widely used in computer vision, speech recognition, natural language processing, audio recognition, bioinformatics and other fields, and has achieved excellent results. Although DL has achieved great success in this century, but can the existing DL really have the intelligence of the human brain and the ability to think like humans? 
In \cite{NeuroLogic}, when given a sentence with disordered words, the human brain can still easily understand the meaning of the sentence, but the DL algorithm is helpless. 
In the study of adversarial attack in \cite{dFooling} , when a person has an attack picture hanging on the abdomen, the classifier cannot recognize human beings, and the judgment is wrong. With the gradual exposure of DL in semantics, image and other fields, the application of DL in the field of wireless communications also encounters many challenges to achieve significantly better performance and more convenient implementations compared to the traditional communication systems, particularly in the complex nonlinear systems \cite{AI-5G}. It is urgent to explore the higher level thinking and cognitive methods of the human brain in artificial intelligence (AI) study for wireless communications and other applications. An important debate in the AI community today is how higher level intelligence and cognition is constituted?

In the wireless communication systems, we noticed that the feedback mechanism has already been deployed in multiple core technologies, including closed-loop large-scale antenna precoding/beamforming technology and interference cancellation technology based on CSI feedback; resource allocation and load balancing technology based on the feedback of resource status and QoS (Quality of Service) variations; channel equalization technology based  on the decision feedback equalizers (DFE), etc. Inspired by the feedback mechanisms and in combination with the existence of human brain's reflective thinking ability, this paper is committed to endowing DL with higher level of intelligence and cognition by exploring the neural networks and algorithms with a feedback structure. Contrasted with the feedforward structure of a system, the conclusions drawn in the integrated information theory (IIT 3.0) \cite{Tononi3} also verifies that an integrated system with feedback is autonomous due to it can take action and respond to its internal state, and therefore has the stronger consciousness.

In the direction of network design with feedback, there have been some research results on it. In wireless communication systems, ISI caused by such as multipath signal propagation, nonlinear signal distortions, imperfect design of wireless transceivers, and channel fading environments, etc., will increase the error rate and weaken the reliability of the communication system. In order to conquer ISI, as early as 1985, Qureshi et al. proposed to use the decision feedback mechanism to improve the performance of the equalization module in the digital signal communication system \cite{I-intro-8}. In the subsequent research, Kim et al. added a discriminative feedback mechanism to the original decision feedback equalizer, which further improved the equalization effect \cite{I-intro-10}. Siu et al. introduced a three-layer perceptron neural network into the structure of the original decision feedback equalizer, and screened and compared the feedback signals. The results show that the performance of the equalizer with the feedback mechanism is better than that without it, and the performance of the equalizer that selects the correct decision signal for feedback is better than that of the equalizer that feeds back all signals \cite{Siu1990}. However, the equalizer cannot guarantee low error rate when the channel nonlinearity is strong and the multipath is abundant. In 5G  and beyond 5G networks, as the propagating signal at mmWave communication systems suffers from high free space propagation loss, atmospheric loss, rain attenuation, and material penetration loss, etc., ISI cancellation becomes a more challenging issue. 

In the image processing field, some interesting networks with feedback have also been presented. Among them, Karim, et al. proposed RIGN which improved the performance of the network by adding a simple canonical loop gate control mechanism to the deep convolutional neural networks (CNN) in \cite{8658560}; Li et al. proposed a ``rethinking" learning algorithm by adding a feedback layer and generating an emphasis vector in \cite{LI2018183}; Nayebi et al. proposed ConvRNNs to explore the role of recursion in improving classification performance in \cite{2018arXiv180700053N}; and Caswell et al. proposed Loopy Neural Net (LNN) to simulate the feedback loop in the brain by unrolling several time steps of the recurrent neural network in \cite{Caswell2016LoopyNN}. The results of these network designs all show that the existence of a feedback structure can effectively improve network performance. However, the design of the existing solutions is mostly based on feedback and analysis of the data of a module or hidden layers, without considering more complex rethinking capabilities of humans or the realization of machine intelligence. Thus, the performance of the algorithm is insufficient. Moreover, the existing feedback schemes basically lack theoretical modeling of the network model, rigorous theoretical analysis and verification, and thus cannot guarantee the effectiveness of the model in complicated applications.

From the above investigation in either wireless communications or image processing fields, in spite of many feedback based DL networks have been applied and achieved better gains, most of the existing solutions adopt a structure with direct data feedback, which lacks intelligent processing of the feedback data, and therefore leads to that the existing feedback structure is unexplainable and too mechanical. This also effects its applications on more complex situations, e.g., complicated nonlinear systems, systems with the increased and nonindependent parameters, and scenarios with serious channel distortion  \cite{AI5G}, etc. Due to lack of the design concept of the feedback modules, the current designs also lack the thinking about the relationship between the feedback module and the feedforward modules, which simulate the collaboration of multiple function modules of the human brain. Finally, the existence of the human brain's ability to rethink is the function of a complex system. The simple information loops of a neural network are not considered to have achieved machine intelligence. It is necessary to consider the information sharing and mutual cooperation among the various modules in the system. 
In this paper, focusing on the integration of the frontier areas of DL and feedback mechanisms, we propose a multi-agent feedback architecture named MAFENN, which mainly targets on the improvement of classical feedforward DL networks.    
The MAFENN framework, consisting of three cooperative agents, namely Encoder, Feedbacker and Processor, as shown in Fig.~\ref{fig_sim}, aims to intelligently remove the noise and recover the clean data as much as possible to facilitate the subsequent downstream tasks. With the introduction of feedback modules in MAFENN, mathematical modeling and the convergence issues of the networks are then addressed to provide a stronger theoretical support to our proposal. Finally, in order to verify our MAFENN framwork, focusing on the wireless equalization, a MAFENN framework based equalizer, i.e. MAFENN-E, is formulated and modeled to solve the equalization problem in wireless multipath channels with either linear or nonlinear signal distortions. Different from the existing DL solutions with simple feedback loops, which lack intelligent rethinking capabilities and theoretical modeling. Based on the two-player game modeling \cite{pmlr-v119-fiez20a}, we mathematically formulate the network as a three-player Stackelberg game and provide theoretical modeling and analysis. As far as we know, our work is the first multi-agent system work to introduce a feedback agent in the feedforward DL networks and provides the theoretical modeling of the network. The key contributions of our paper are summarized as follows.
\begin{enumerate}
    \item A noval multi-agent MAFENN framework is proposed, which consists of three fully cooperative agents to solve the problem that the conventional feedback mechanism lacks feedback learning ability from feedback data and makes DL networks more intelligent. 
    \item The MAFENN framework is mathematically formulated as a three-player Feedback Stackelberg game, and this  game is proved to converge to the Feedback Stackelberg equilibrium. 
    \item For wireless channel equalization, the issue is formulated as a conditional probability distribution learning and feedback learning problems, and MAFENN-E is proposed for wireless communications with ISI interference. Simulation results show the effectiveness and  robustness of our proposal in either linear or nonlinear channels. 
    
\end{enumerate}

In this section, we firstly introduce the motivation of our MAFENN proposal, survey the feedback related network literature in both wireless communication and image processing fields, and briefly address our work and our contributions. In Section~\ref{sec:new_framework}, in terms of the general neural network innovation, we mainly introduce the details of the structure of our multi-agent feedback based neural network, i.e., MAFENN, Feedback Stackelberg game formulation, Stackelberg learning dynamics and convergence proof.
As a general framework, MAFENN can be applied to promote the solutions which are developed based on traditional feedback mechanism or feedforward networks in many fields, e.g., intelligent communications.
In Section~\ref{sec:equalizer}, we mainly explore the application of MAFENN on wireless equalizers. In order to help readers learn the state of arts of wireless equalizers, particularly machine learning enabled wireless equalizers, we further introduce some related works for wireless equalization. Then formulate the problem and propose the MAFENN-E network model to find the optimal solutions. Subsequently, the training and testing results of  MAFENN-E and other counterpart equalizers on the same training and testing dataset are shown in Section~\ref{sec:simulation}. Finally, the conclusion is drawn in Section~\ref{sec:conclusion}.

\section{MAFENN framework}
\label{sec:new_framework}
In this section, we describe the main methods of our proposed MAFENN framework.
Firstly, the details of the feedback learning model is presented in Section~\ref{network model}. Secondly, the feedback learning model is formulated as a multi-agent feedback Stackelberg game in Section~\ref{Method_game}. Then the learning dynamics of MAFENN is analyzed in Section~\ref{Method_learningdynamics}. Finally, the convergence and training process is discussed in Section~\ref{sec:convergence analysis}.

\subsection{Feedback Learning Model}
\label{network model}

\begin{figure}[htbp] 
\centering
\includegraphics[width=3.5 in]{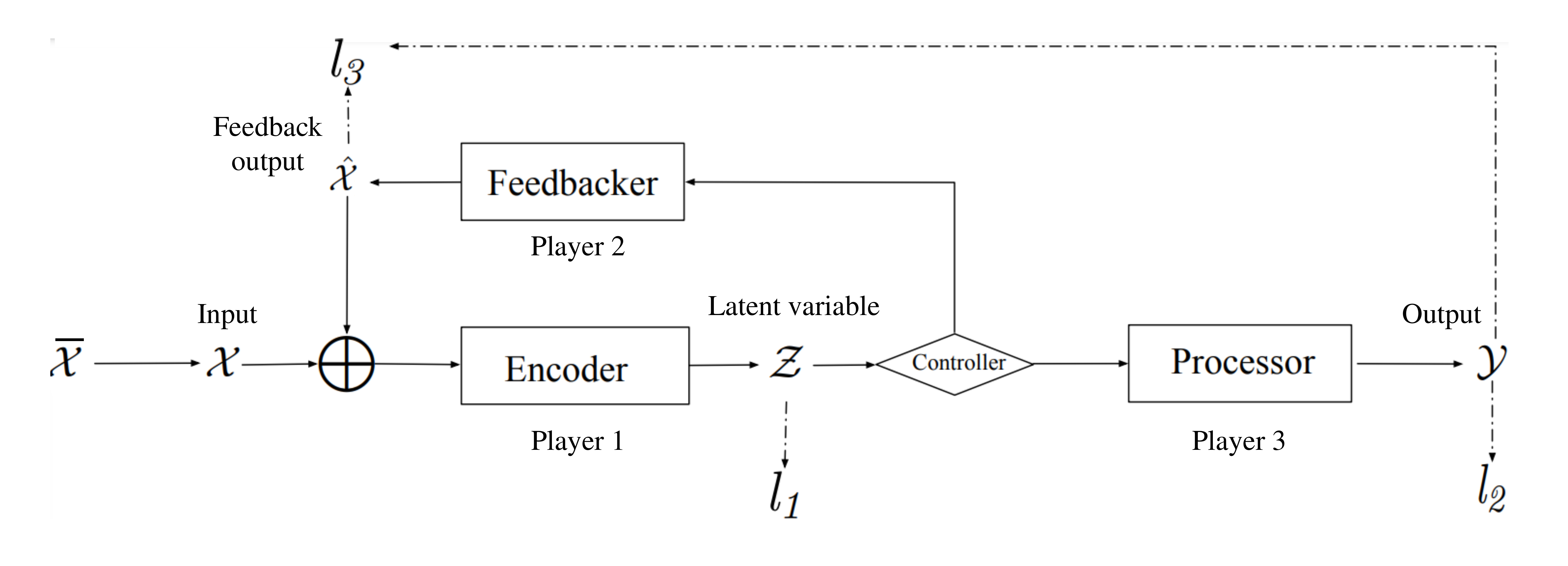}
\caption{Network structure of MAFENN and schematic diagram of Feedback Stackelberge game}
\label{fig_sim}
\end{figure}

Based on a general feedforward DL network structure with an encoder and a processor for the downstream task implementation, in order to simulate the feedback thinking capability in DL, we aim to add a feedback learning module and propose a MAFENN framework, to give the entire system stronger feedback learning skill and more intelligent information processing power.  Fig.\ref{fig_sim} shows the network structure of MAFENN which consists of three cooperative agents, i.e., Encoder, Feedbacker and Processor.

In this section, the notations used for the MAFENN representation will be introduced firstly. We consider a dataset with the generated $i.i.d$ data from the distribution $\mathcal{D}$ over $\mathcal{X}\times\mathcal{Y}$, where $\mathcal{X}$ is the input space and $\mathcal{Y}$ is the label space. The proposed MAFENN framework consists of three agents as shown in Fig.~\ref{fig_sim}. The Encoder agent $E(\cdot)$ that encodes the input ${x} \in \mathcal{{X}}$ and feedback corrected data $\hat{x} \in \hat{\mathcal{X}}$ to a latent variable $z = E({x},\hat x)$, where $z\in \mathcal{Z}$. The put of Encoder could be either a concatenation of $\hat x$ and ${x}$ or a substitution of $\hat x$ to ${x}$. For clear description in the paper, we mainly consider the concatenation method in the following sections. Detailed analysis of these two methods will be discussed in Section~\ref{Sec:Hyperparameter Optimization}.  The Feedbacker agent $F(\cdot)$ that maps the latent variable $z$ to the feedback corrected data $\hat{x}=F(z)$ where $\hat{x}\in \hat{\mathcal{X}}$. The Processor network $P(\cdot)$ that produces the label output $y=P(z)$. Controller determines the number of feedback cycles.

In feedforward DL networks, the Stackelberg game can be formalized as a bilevel optimization problem in which the upper level optimization problem is concerned with minimizing the learner’s cost post adversarial data transformation, while the lower level optimization problem of finding the optimal data transformation becomes the constraint of the upper level problem. The theoretical definition and discussions on the convergence issues has been studied in \cite{pmlr-v119-fiez20a}. In this paper, based on the Stackelberg game, we further formulate our MAFENN framework as a Feedback Stackelberg game. Considering the case of three cooperative players, let $N=\{\rm{Player~1},\rm{Player ~2},\rm{Player ~3}\}$ be the set of players. For each $i\in N$, let $A_i$ be the corresponding set of actions. Without loss of generality, Encoder $E(\cdot)$ can be denoted as Player 1 and deemed as leader of Player 2 and Player 3. Feedbacker $F(\cdot)$ is Player 2, and Processor $P(\cdot)$ is Player 3, which are followers. In our MAFENN framework, the action set $A =\{\mathcal{X}_1,\mathcal{ X}_2,\mathcal{X}_3 \}= \{\mathcal{Z},\mathcal{Y},\mathcal{\hat X}\}$. Let $F=\{\l_1,\l_2,\l_3\}$ be the set of cost functions for the corresponding players. We assume $\l_2\propto  \l_3$, they are positively correlated, and $\l_1=\l_2+\l_3$. Algorithm~\ref{alg: forward} shows the pseudocode of the MAFENN process.


\begin{algorithm}[htb]
    \caption{The algorithm of MAFENN}
    \label{alg: forward}
    \begin{algorithmic}[1]
    \STATE {\bfseries Input:} Encoder network $E{(\cdot)}$, Feedbacker network \\ $F{(\cdot)}$, Processor network $P{(\cdot)}$, the length of the signal processing window $N$, the length of the feedback learning window $K\ (K\geq 0)$, the number of the \\ feedback cycles $C\ (C\geq 0)$, the concatenation operation $op$, and the length of the number of transmitted data $L\ (L\geq 0)$.
    \FOR{i = 1 to L}
        \STATE Collect the past received signal ${{x}}^{(i)}_{raw}=[{{x}}^{(i)}, {{x}}^{(i-1)}, \dots, {{x}}^{(i-N+1)}]^T$, if $i-k<0$, \\ let $x^{(i-k)}=0$.
        \STATE Collect the past feedback signal $\hat x^{(i)}_{feed}=[x^{(i)}, \hat x^{(i-1)}, \dots, \hat x^{(i-K)}]^T$, if $i-k<0$, let $\hat x^{(i-k)}=0$.
        \FOR{j = 1 to C}
            \STATE \quad $ z_{mid} = E(op(\hat{x}^{(j-1)}_{feed},{{x}}^{(i)}_{raw}))$, 
            \STATE \quad $\hat{x}^{(j)} = F(z_{mid})$ 
        \ENDFOR
        \STATE $\hat x^{(i)}_{feed}=[\hat{x}^{(C)}, \hat x^{(i-1)}, \dots, \hat x^{(i-K)}]^T$
        \STATE $z = E(op(\hat{x}^{(i)}_{feed},x^{(i)}_{raw}))$
        \STATE ${y} = P(z)$ 
    \ENDFOR
  \end{algorithmic}
\end{algorithm}  

\subsection{Feedback Stackelberg Game Model}
\label{Method_game}

Before we formulate the proposed MAFENN, we first define the Feedback Stackelberg game, which is a three-player Stackelberg game.
Fig. \ref{fig_sim} also shows the schematic diagram of the Feedback Stackelberg game.
Consider a Stackelberg game between three agents where Player 1 is deemed the leader to Player 2 and Player 3 , Player 2 is the leader to Player 3 but the follower to Player 1, and Player 3 is the follower to Player 1 and Player 2. 
Following the definition of the n-player differentiable game \cite{balduzzi2018mechanics}, we define the Feedback Stackelberg game as follows.
\begin{definition}
    \label{FeedbackStackelberggame}
    \textbf{(Feedback Stackelberg Game).}
    A Feedback Stackelberg game is a special three-player Stackelberg game with a feedback structure, as shown in Fig. \ref{fig_sim}, which can be defined to be the tuple $(3, \mathcal{X}_1, \mathcal{X}_2, \mathcal{X}_3, l_1, l_2, l_3)$, where for all $i\in \{1,2,3\}$, $\mathcal{X}_i \subseteq \mathbb{R}^{m_i}$ is the parameters (actions) for Player $i$, $l_i:\mathbb{R}^m \rightarrow \mathbb{R}$  is a twice continuously differentiable objective function of all the parameters, where $m=\sum_{i=1}^3 m_i$.
\end{definition}

Furthermore, we can define our proposed unified model as a Feedback Stackelberg game in the form of the tuple $(3,x_1,x_2,x_3,l_1,l_2,l_3)$, where for all player $i$ such that $1\leq i \leq 3, x_i\in \mathbb{R}^{m_i}$, and each Player $i$ has a loss function $l_i:\mathcal{X}\rightarrow \mathbb{R}$, where $\mathcal{X}=\mathcal{X}_1\times \mathcal{X}_2\times \mathcal{X}_3\in \mathbb{R}^m$ with $\mathcal{X}_i \in \mathbb{R}^{m_i}$, $\sum_i^3 m_i=m$ denoting the action spaces of the three players. The three players aim to solve the following optimization problems, respectively:
\begin{equation}
    \label{optimization}
    \begin{aligned}
    \min_{x_1\in \mathcal{X}_1}\quad &l_1(x_1,x_2,x_3), \\
    \text{s.t.}\quad &x_2\in \arg \min_{y\in \mathcal{X}_2} l_2(x_1,y,x_3),\\
    &\text{s.t.}\quad x_3\in \arg \min_{z\in \mathcal{X}_3} l_3(x_1,x_2,z)
    \end{aligned}
\end{equation}
where $\mathcal{X}_1\in \mathbb{R}^{m_1}, \quad\mathcal{X}_2\in \mathbb{R}^{m_2}, \quad \rm{and} \quad  \mathcal{X}_3\in \mathbb{R}^{m_3}$ denote the parameter spaces of Encoder, Processor and Feedbacker.

Then we discuss the concept of the Feedback Stackelberg equilibrium. Considering the generality of our proposed framework, it can also be applied to non-convex or non-concave target functions.
We only focus our attention on local notions of the equilibrium concepts. 
Therefore, we have the local notion of the three-player Feedback Stackelberg equilibrium.
\begin{definition}
\label{LFSE_def}
\textbf{(Local Feedback Stackelberg Equilibrium (LFSE)).}
    Consider $U_i\cap X_i$ for each $i\in\{1,2,3\}$. The strategy $x_1^*\in U_1$ is a local Stacelberg solution for the leader if, $\forall x_1\in U_1$,
    \resizebox{.99\hsize}{!}{
    $
    \sup_{\begin{aligned}
        &x_2\in R_{U_2}(x_1^*),\\
        &x_3\in R_{U_3}(x_1^*,x_2)
    \end{aligned}}l_1(x_1^*,x_2,x_3)\leq  \sup_{\begin{aligned}
        &x_2\in R_{U_2}(x_1),\\
        &x_3\in R_{U_3}(x_1,x_2)
    \end{aligned}}l_1(x_1,x_2,x_3)
    $
    },
    where $R_{U_2}(x_1)=\{y\in U_2|l_2(x_1,y,x_3)\leq l_2(x_1,x_2,x_3)$, $x_3\in R_{U_3}(x_1,x_2),\forall x_2\in U_2\}$, and $R_{U_3}(x_3)=\{z\in U_3|l_3(x_1,x_2,z)\leq l_3(x_1,x_2,x_3), \forall x_3\in U_3\}$. That means $(x_1^*,x_2^*,x_3^*)$ for any $x_2^*\in R_{U_3}(x_1^*),~x_3^*\in R_{U_3}(x_1^*,x_2^*)$ is a local Stackelberg equilibrium on $U_1\times U_2\times U_3$.
\end{definition}

We denote $D_il_i$ as the derivative of $l_i$ with respect to $x_i$, $D_{ij}l_i$ as the partial derivative of $D_il_i$ with respect to $x_j$.
$\hat{\mathcal{D}}_1 l_1$ denoting the derivative of $l_1$ with respect to $x_1,x_2=r(x_1),x_3=h(x_1,r(x_1))$, i.e., $\hat{\mathcal{D}}_1 l_1 (x_1,x_2,x_3)=D_1l_1+D_2l_1 D_{x_1}x_2+D_3l_1 D_{x_1}x_3.$ 
$\hat{\mathcal{D}}_2 l_2$ denoting the derivative of $l_2$ with respect to $x_2, x_3=h(x_1,r(x_1))$, i.e., $\hat{\mathcal{D}}_2 l_2 (x_2,x_3)= D_2l_2+D_3l_2D_{x_2}x_3$.
$\hat{\mathcal{D}}_3 l_3=D_3l_3$ denotes the derivative of $l_3$ with respect to $x_3$.
Then, the following definition gives the sufficient conditions for a LFSE.
\begin{definition}
    \label{DFSE_def}
\textbf{(Differential Feedback Stackelberg Equilibrium (DFSE)).}
    The joint strategy $x^*=(x_1^*,x_2^*,x_3^*)\in X$ is a differential Stackelberg equilibrium if $\hat{\mathcal{D}}_1 l_1(x^*)=0,\hat{\mathcal{D}}_{2}l_2(x^*)=0, \hat{\mathcal{D}}_3 l_3(x^*)=0$, $\hat{\mathcal{D}}^2_1 l_1(x^*)>0$, $\hat{\mathcal{D}}^2_{2}l_2(x^*)>0$ and $\hat{\mathcal{D}}_3^2l_3(x^*)>0$, where $x_2^*=r(x_1^*), x_3^*=h(x_1^*,x_2^*)$, and $r(\cdot)$ and $h(\cdot)$ implicitly defined by $\hat{\mathcal{D}}_{2}l_2(x^*)=0,~  \hat{\mathcal{D}}_3 l_3(x^*)=0$.
\end{definition}

\subsection{Learning Dynamics of MAFENN}
\label{Method_learningdynamics}

In this section, we firstly design a novel neural network structure with a feedback agent on a feedforward network in Section \ref{network model}. In order to give mathematical analyses, we further formulate the whole structure into a three-player Stackelberg game and identify the optimization target in Section \ref{Method_game}. Subsequently, Local Feedback Stackelberg Equilibrium (LFSE) and Differential Feedback Stackelberg Equilibrium (DFSE) of the three-player Stackelberg game in Definition~\ref{LFSE_def} and Definition~\ref{DFSE_def} are defined, respectively. If the differential conditions in Definition~\ref{DFSE_def} are satisfied, DFSE is equivalent to LFSE. 
In \cite{pmlr-v119-fiez20a} for two-player Stackelberg game,
gradient-based learning dynamics were derived to emulate the natural structure of a two-player Stackelberg game using the implicit function theorem and further the convergence of the implicit learning dynamics is proved. 
For our proposed network modeled as three-player Stackelberg game, as enlightened by \cite{pmlr-v119-fiez20a}, the learning dynamics for the three-player Stackelberg game are given to solve DFSE in this subsection. Then the differomorphism and convergence are proven in Section \ref{sec:convergence analysis}.

Let $\omega(x)=(\hat{\mathcal{D}}_1 l_1,\hat{\mathcal{D}}_2 l_2,\hat{\mathcal{D}}_3 l_3)$ be the gradient vector for the Feedback Stackelberg game. $\omega_i$ denotes the $i$-th element of $\omega$, i.e., the gradient of $i$-th player.
The derivative of Player 1 is 
\begin{equation}
\resizebox{0.95\hsize}{!}{
$
\begin{aligned}
\omega_1 =& D_1 l_1(x_k) + D_2 l_1(x_k) D_1 r(x_k) \\
&+ D_3 l_1(x_k) \left(D_1 h(x_k) +  D_2 h(x_k) D_1 r(x_k) \right),
\end{aligned}
$
}
\end{equation}
where $r(\cdot),~h(\cdot)$ defined by $\omega_2=0$ with $det(\hat{\mathcal{D}}_2^2l_2(x))\neq 0$, and $\omega_3=0$ with $det(\hat{\mathcal{D}}_3^2l_3(x))\neq 0$.
\begin{equation}
\resizebox{0.95\hsize}{!}{
$
\begin{aligned}
    D_1h(x_k)=&-D_{31}^Tl_3(x_k)(D_3^2 l_3(x_k))^{-1},\\ D_2h(x_k)=&-D_{32}^Tl_3(x_k)(D_3^2 l_3(x_k))^{-1}, \\
    D_1r(x_k)=& -(D_{21}l_2(x_k)+D_{23}l_2(x_k)D_1h(x_k))^T(D^2_2l_2(x_k))^{-1}.
\end{aligned}
$
}
\end{equation}
The derivative of Player 2 is 
\begin{equation}
\omega_2= D_2l_2(x_k)+D_3l_2(x_k)D_2h(x_k).
\end{equation}
The derivative of Player 3 is 
\begin{equation}
\omega_3=D_3l_3(x_k).
\end{equation}
So the Stackelberg learning rule we study for each player is given by
\begin{equation}
    x_{k+1,i} = x_{k,i} -\lambda_{i}\omega_i(x_k).
\end{equation}
In Algorithm~\ref{algo}, we show the pseudocode of the Stackelberg learning dynamics. 

\begin{algorithm}
   \caption{Feedback Stackelberg Learning Dynamics}
    \label{algo}
\begin{algorithmic}[1]
   \STATE {\bfseries Input:} $x^0=(x_1^0,x_2^0,x_3^0)\in X, where\ x_1^0, x_2^0$ are pretrained and $x_3^0$ is initialized randomly. Learning rate $\lambda_3 > \lambda_2 > \lambda_1>0$. 
   \FOR{$k=0,1,\cdots$}
    \STATE{$D_1h(x_k)=-D_{31}^Tl_3(x_k)(D_3^2 l_3(x_k))^{-1}$}
    \STATE{$D_2h(x_k)=-D_{32}^Tl_3(x_k)(D_3^2 l_3(x_k))^{-1}$}
    \STATE{$\begin{aligned}
        D_1r(x_k) =& -(D_{21}l_2(x_k)+D_{23}l_2(x_k)D_1h(x_k))^T\\&
        (D^2_2l_2(x_k))^{-1}
    \end{aligned}$}
    \STATE{
    $
    \begin{aligned}
    \omega_1\leftarrow &D_1 l_1(x_k) + D_2 l_1(x_k) D_1 r(x_k) \\&+ D_3 l_1(x_k)(D_1 h(x_k) + D_2 h(x_k) D_1 r(x_k)
    \end{aligned}
    $
    }
    \STATE $\omega_2\leftarrow D_2l_2(x_k)+D_3l_2(x_k)D_2h(x_k)$
    \STATE $\omega_3\leftarrow D_3l_3(x_k)$
    \STATE $x_{k+1,i}\leftarrow x_{k,i}-\lambda_1\omega_{1}$
    \STATE $x_{k+1,i}\leftarrow x_{k,2}-\lambda_2\omega_{2}$
    \STATE $x_{k+1,i}\leftarrow x_{k,3}-\lambda_3\omega_{3}$
   \ENDFOR
\end{algorithmic}
\end{algorithm}

\subsection{Convergence Analysis}
\label{sec:convergence analysis}
In this section, we perform the convergence analysis and prove that our proposed MAFENN model can reach the fixed point and almost surely avoid the saddle points. The proof of this section is enlightened by \cite{pmlr-v119-fiez20a}, which proves the convergence issue of the two-player Stackelber game.

Recall the Feedback Stackelberg learning rule for each player and rewrite the rule as 
\begin{equation}
    \begin{aligned}
    x_{k+1,1}=&x_{k,1}-\lambda_1 (D_1l_1(x_k)+D_2l_1(x_k)D_1 r(x_k)\\
    &+D_3l_1(x_k)(D_1h(x_k)+D_2h(x_k)D_1 r(x_k))),\\
    x_{k+1,2}=&x_{k,2}-\lambda_2( D_2l_2(x_k)+D_3l_2(x_k)D_2h(x_k)),\\
    x_{k+1,3}=&x_{k,3}-\lambda_3 D_3l_3(x_k).
    \end{aligned}
\end{equation}

Note that by the implicit function theorem,
\begin{equation}
\resizebox{.99\hsize}{!}{
$
\begin{aligned}
D_1h(x_k)&=-D_{31}^Tl_3(x_k)(D_3^2 l_3(x_k))^{-1},\\
D_2h(x_k)&=-D_{32}^Tl_3(x_k)(D_3^2 l_3(x_k))^{-1},\\
D_1r(x_k) &= -(D_{21}l_2(x_k)+D_{23}l_2(x_k)D_1h(x_k))^T(D^2_2l_2(x_k))^{-1}.
\end{aligned}
$
}
\end{equation}

The above Stackelberg update is equivalent to the dynamics:
\begin{equation}
    \begin{aligned}
    x_{k+1,1}=&x_{k,1}-\frac{\lambda_3}{\tau_1} (D_1l_1(x_k)+D_2l_1(x_k) D_1 r(x_k)\\
    &+D_3l_1(x_k)\left(D_1h(x_k)+D_2h(x_k)D_1 r(x_k)\right)),\\
    x_{k+1,2}=&x_{k,2}-\frac{\lambda_3}{\tau_2}( D_2l_2(x_k)+D_3l_2(x_k)D_2h(x_k)),\\
    x_{k+1,3}=&x_{k,3}-\lambda_3 D_3l_3(x_k),
    \end{aligned}
\end{equation}
where $\tau_1=\frac{\lambda_3}{\lambda_1},\tau_2=\frac{\lambda_3}{\lambda_2}$ are the ``timescale" separation. So we can write the Stackelberg update in ``vector" form as:
\begin{equation}
x_{k+1}=x_k-\lambda_3\mathit{\omega}(x_k),
\end{equation}
where
\begin{equation}
    \mathit{\omega}(x_k)=(
    \tau_1^{-1}\hat{\mathcal{D}}_1l_1(x_k),
    \tau_2^{-1}\hat{\mathcal{D}}_2l_2(x_k),
    \hat{\mathcal{D}}_3l_3(x_k)).
\end{equation}
So the update is equivalent to 
\begin{equation}
    g(x)=x-\lambda_3(
    \tau_1^{-1}\hat{\mathcal{D}}_1l_1(x),
    \tau_2^{-1}\hat{\mathcal{D}}_2l_2(x),
    \hat{\mathcal{D}}_3l_3(x)
    ).
\end{equation}

Then we can have the next results, which shows the unified Feedback Stackelberg game can avoid saddle points almost surely under our assumptions.
\begin{corollary}[Almost Sure Avoidance of Saddle Points.] 
\label{thm:avoid-saddle}
Consider a general Feedback Stackelberg game defined by $(3,x_1,x_2,x_3,l_1,l_2,l_3)$, where $l_i:\mathbb{R}^m \rightarrow \mathbb{R}$ is twice differentiable loss. Player 1 is the leader to Player 2 and Player 3; Player 2 is the follower to Player 1 and leader to Player 3; and Player 3 is the follower to Player 1 and Player 2. Suppose that $\omega$ is $L$-$Lipschitz$ with $\tau_1>1,\tau_2>1$ and that $\lambda_3<1/L$. The Stackelberg learning dynamics converge to saddle points of $\dot{x}=-\omega(x)$ on a set of measure zero.
\end{corollary}

To prove Corollary~\ref{thm:avoid-saddle}, we first show that the update rule $g$ is a diffeomorphism.
\begin{lemma}
Consider a game defined by $(3,x_1,x_2,x_3,l_1,l_2)$, where $l_i:\mathbb{R}^m \rightarrow \mathbb{R}$ is twice differentiable loss. Suppose that $\omega$ is $L$-$Lipschitz$ with $\lambda_3<1/L$. The Stackelberg update $g(x)=x-\lambda_3(
    \tau_1^{-1}\hat{\mathcal{D}}_1l_1(x),
    \tau_2^{-1}\hat{\mathcal{D}}_2l_2(x),
    \hat{\mathcal{D}}_3l_3(x)
    )$, where $\tau_1>\tau_2>1$ and $\tau_1=\frac{\lambda_3}{\lambda_1},\tau_2=\frac{\lambda_3}{\lambda_2}$. The Stackelberg update $g$ is a diffeomorphism.
\end{lemma}

\textbf{Proof. }
The game Jacobian for the Stackelberg update is given by:
\begin{equation}
    \resizebox{.99\hsize}{!}{$
    \displaystyle
    J=\begin{bmatrix}
    \frac{1}{\tau_1} D_1(\hat{\mathcal{D}}_1l_1(x)) & \frac{1}{\tau_1} D_2(\hat{\mathcal{D}}_1l_1(x)) &\frac{1}{\tau_1} D_3(\hat{\mathcal{D}}_1l_1(x))\\
    \frac{1}{\tau_2} D_1(\hat{\mathcal{D}}_2l_2(x)) &\frac{1}{\tau_2} D_2(\hat{\mathcal{D}}_2l_2(x)) &\frac{1}{\tau_2} D_3(\hat{\mathcal{D}}_2l_2(x)) \\
    D_1(\hat{\mathcal{D}}_3l_3(x)) & D_2(\hat{\mathcal{D}}_3 l_3(x)) & D_{3}(\hat{\mathcal{D}}_3l_3(x)) &
    \end{bmatrix}.
    $
    }
\end{equation}

Now, observe that $Dg=\mathit{I}-\lambda_3J(x)$. Then, denote $\rho(A)$ as the spectral radius of a matrix A, and we know that $\rho(A)\leq \|A\|$ for any square matrix A. We also have a assumption $\sup_{x\in \mathbb{R}^m}\|J(x)\|_2\leq L < \infty$, which implies that $\omega$ satisfies the Lipschiz condition on $\mathbb{R}^m$.
Therefore,
\begin{equation}
\begin{aligned}
    \rho(\lambda_3J(x))&\leq \|\lambda_3J(x)\|_2\\
    &\leq \lambda_3 \sup_{x\in \mathbb{R}^m}\|J(x)\|_2\\
    &\leq \lambda_3 L\\
    &<1.
\end{aligned}
\end{equation}
Due to the spectral radius is the maximum absolute value of the eigenvalues,  the above derivation implies that all eigenvalues of $\lambda_3 J(x)$ have an absolute value less than 1. 
So $D_g$ is invertible, the implicit function theorem \cite{lee2013smooth} implies that $g$ is a local diffeomorphism.

\begin{lemma}
    \label{injective_surjective}
    Given a small enough $\lambda$, and $g(x)\triangleq x+\lambda \omega(x)$ is a map $\mathbb{R}^m\rightarrow \mathbb{R}^m$, where $\omega$ is a L-Lipschitz map with $\lambda L \leq \frac{1}{3}$ from $\mathbb{R}^m$ to $\mathbb{R}^m$. Then $g$ is a injective and surjective map from $\mathbb{R}^m\rightarrow \mathbb{R}^m$.
\end{lemma}

\textbf{Proof.} Firstly, we prove that $g$ is a injective map.
Consider $x\neq y$ and suppose $g(x)=g(y)$ so that $y-x=\lambda_3(\omega(y)-\omega(x))$.

According to the assumption $\omega$ satisfies the Lipschiz condition on $\mathit{R}^m$, we have $\|\omega(y)-\omega(x)\|_2\leq L\|y-x\|_2$. Then $\|x-y\|_2\leq L \lambda_3 \|y-x\|_2< \|y-x\|_2$ due to $\|x-y\|_2\leq  \|\lambda_3 (\omega (y)-\omega (x)\|_2$. 
It goes against our assumption. So $g$ is injective.

Then we show that $g$ is a surjective map. We only need to prove that for $\mathit{R}$ large enough, $B_{R^{(0)}}\subset g(B_{2R^{(0)}})$.

Let $\|\omega(0)\|=A$, and we have $\|\omega(x)-\omega(y)\|\leq L\|x-y\|$. Then 
    \begin{align}
        \|g(x)\|\geq& \|g(x)-g(0)\|-\|g(0)\| \\
        =&\|x+\lambda(\omega(x)-\omega(0))\|-\|g(0)\| \\
        \geq & \|x\|-\lambda\|\omega(x)-\omega(0)\|-\|\omega(0)\| \\
        \geq & (1-\lambda L)\|x\|-A \\
        \geq & \frac{2}{3}\|x\|-A \label{surjectivity_1}\\
        \geq & \frac{1}{2} \|x\| \label{surjectivity_2}.
    \end{align}
The line (\ref{surjectivity_1})  is under the assumption $\lambda_3 L\leq \frac{1}{3}$ and the line (\ref{surjectivity_2}) is under the assumption $\|x\|\geq 6A$. Donate $\partial B_{2R}$ as the boundary of $B_{2R}$, and we have $g(\partial B_{2R})\subset \mathbb{R}^n \setminus B_R$.

We assert that $B_R \subset g(B_{2R})$. If $B_R \subset g(B_{2R})$ is not true, then there exists $x_0\in B^0_R, x_0 \notin g(B_{2R})$ due to $g$ is a continuous map. $B_{2R}$ and $g(B_{2R})$ are closed set, so there exists $\epsilon \geq 0$ such that $B_\epsilon(x_0)\not\subset g(B_{2R})$.

Then we can define $\phi:B_{2R}\rightarrow \partial B_R$. 
If $\|g(x)\|<R, \phi(x)=y,$ we have $ y=x_0+\alpha(g(x)-x_0)$ for $\alpha>0$. So $\|y\|=R$. 
If $\|g(x)\|\geq R$, $\phi(x)=\frac{g(x)}{\|g(x)\|} R$. Then $\phi$ is a continuous map from $B_{2R}\rightarrow \partial B_R$ and the degree of $\phi:B_{2R}\rightarrow \partial B_R$ is 1. This is because $\phi |_{\partial B_{2R}}$ is homotopic to $\frac{x}{\|x\|}R$.
However, it is impossible because the assumption $x_0 \notin g(B_{2R})$ is illegal.
So $B_R \subset g(B_{2R})$, and $g$ is a surjective map. 

Therefore, the inverse of $g$ is well-defined and since $g$ is a local diffeomorphism on $\mathbb{R}^m$. $G^{-1}$ is smooth on $\mathbb{R}^m$. Thus, $g$ is a diffeomorphism.

Remark the proof in \cite{pmlr-v119-fiez20a}, which shows that the set of initial points that finally converges to the saddle points has measure zero. Therefore, we have proved that the Stackelberg learning dynamics converge to saddle points of $\dot{x}=-\omega(x)$ on a set of measure zero.$\hfill\blacksquare$

\section{MAFENN framework for channel equalization}
\label{sec:equalizer}

As we all know that the main target of wireless communication systems is to provide high efficiency and reliable signal transmissions. With the rapid development of ML in the recent years, one way to improve the ML enabled wireless communications is to rely on priori models to enhance the interpretability of the neural network models and abstract more accuracy information from the systems  \cite{GrayBox}. In this paper, focused on DL enabled wireless communication modules which can originally be realized by feedforward structure DL neural networks, such as source coding/decoding, channel coding/decoding, channel equalization, resource allocations, the MAFENN framework is proposed to provide high-level intelligence with an additional feedback learning agent.  MAFENN is designed with additional capability to rethink the high-level features and improve the noise cancellation or error correcting. This capability is really suitable for DL-based solutions for wireless communications to further improve signal recovery and transmission performance. In addition, the multi-agent structure can solve complicated issues with downstream tasks after rethink processing, such as decoding after signal recovery in equalization to be illustrated in this section. The possible other applications include source/channel coding/decoding, UE section, channel allocation, MCS selection and beam selection, which can be abstracted as classification problems. More possibilities will be explored in our future studies and we believe that our proposed algorithm can be applied to more wireless communication issues. 

In this section,  we  propose a novel network based on the MAFENN framework for equalizer to overcome ISI in wireless communications and further improve the decoding success probability. We firstly introduce the current challenges of the state-of-art equalization methods in Section~\ref{sec:MAFENN equalizer challenges}. Then formulate the adaptive channel equalization problem to a conditional probability distribution learning in Section~\ref{sec:MAFENN equalizer formulation}. Finally, the detailed components of the MAFENN framework based equalizer are described in Section~\ref{sec:MAFENN equalizer components}. 

\subsection{Related Work and Challenges}
\label{sec:MAFENN equalizer challenges}

Before DL was widely used, conventional equalization methods were usually divided into two categories, i.e., linear and nonlinear equalization. Zero-forcing equalizers (ZF), Minimum mean-square error equalizers (MMSE), least mean square equalizer (LMS) and recursive least squares  equalizer (RLS) \cite{MMSE}\cite{RLS} are commonly used in the linear equalization. However, linear equalization cannot solve the problem well when channel distortion is severe. Nonlinear equalizers are usually proposed like the maximum likelihood symbol detection (MLSD) assisted equalizer, the maximum likelihood sequence estimation (MLSE) \cite{MLSE} assisted equalizer and DFE, as mentioned in \cite{I-intro-8}. When the channel has a deep spectral null in its bandwidth, the performance of linear equalization will be very poor because the equalizer will set a high gain at the frequency of the spectral null, which will enhance the additional noise in the frequency band. DFE was proposed to overcome this limitation. It consists of two parts: the forward filter and the feedback filter, which counteract the ISI caused by the previous symbols and post symbols respectively. Li et al. further proposed an adaptive decision feedback equalizer using error feedback, which not only improves the performance, but also weakens the error propagation in  \cite{II-B-1-3}. In recent years, to further conquer the  nonlinearity in high-speed channels and the large delay spread in the underwater acoustic channels, the enhancement DFE based equalization methods are proposed in \cite{DFE2020} and \cite{DFEwater}. 

In the traditional methods, for either linear or nonlinear methods, the complexity of equalization depends on the channels and may be very high in practical
communication environments, where multiple reflections, multiple refractions, scattering, etc., exist. In these environments, the traditional equalizers have huge computational cost, e.g., tracking complex CSI and matrix operations to update the fitter parameters. Moreover, the performance of the traditional methods also decrease in complex tasks such as nonlinear systems, systems with increased parameters and parameters are not independent. With the rapid development of ML in wireless communication fields,  ML-based equalizers were proposed for ISI suppression and signal distortion for the simplicity of modeling and strong fitting ability. Gibson et al. proposed an adaptive equalizer using a neural network architecture based on the multilayer perception (MLP) to combat ISI in linear channels with white Gaussian noise~\cite{I-intro-2}. Following this work, Gibson et al. further applied the MLP-based equalizer to nonlinear channels with colored Gaussian noise, which demonstrated that the bit error rate (BER) performance of  MLP-based equalizer is  close to that of the optimal equalizer~\cite{I-intro-5}. With the aid of DL, Ye et al. jointly designed channel equalization and decoding, and demonstrated the  robust performance under various channel conditions, including a time-varying frequency selective channel that generates severe ISI~\cite{I-intro-6}. Following this work, in~\cite{I-intro-7}, the DL neural network structure was utilized to recover data symbols conveyed in OFDM principles. When the feedback mechanism is considered in DL, many feedback-based ML algorithms were proposed to solve the channel equalization problem~\cite{I-intro-8}-\cite{Siu1990}. The main ideas of them have briefly been introduced in Section~\ref{introduction}. As we know, a recurrent neural network (RNN) can ideally implement the inverse of a finite memory system, with the result that it can substantially model a nonlinear infinite memory filter in principle. Therefore, some RNN-based equalizers, e.g.,  \cite{Kechriotis1994}, \cite{Xiaoqin2004} and \cite{Park2008},  were proposed to solve equalization problems. In RNN, feedback loops are involved to represent sequence processing of different samples.  However, for each sample data input, no additional feedback processing is involved to introduce additional learnable agents to reprocess the disturbed data and improve the feature expression.

Although many methods based on feedback mechanisms assisted ML algorithms have been applied in the field of equalization, most of these methods only use the structural characteristics of the feedback mechanism to leverage the knowledge of wireless channel conditions and the passed information. These methods lack enough intelligent learning ability from feedback data, thus the performance improvement needs to be further explored in complex situations. Our proposed MAFENN framework has stronger feedback capabilities and more intelligence, which can effectively overcome these problems. So MAFENN-E is proposed to conquer wireless fading channels with ISI.

\subsection{Problem Formulation and Optimization}
\label{sec:MAFENN equalizer formulation}

\begin{figure*}[htbp]
    \centering
    \includegraphics[width=0.95\textwidth]{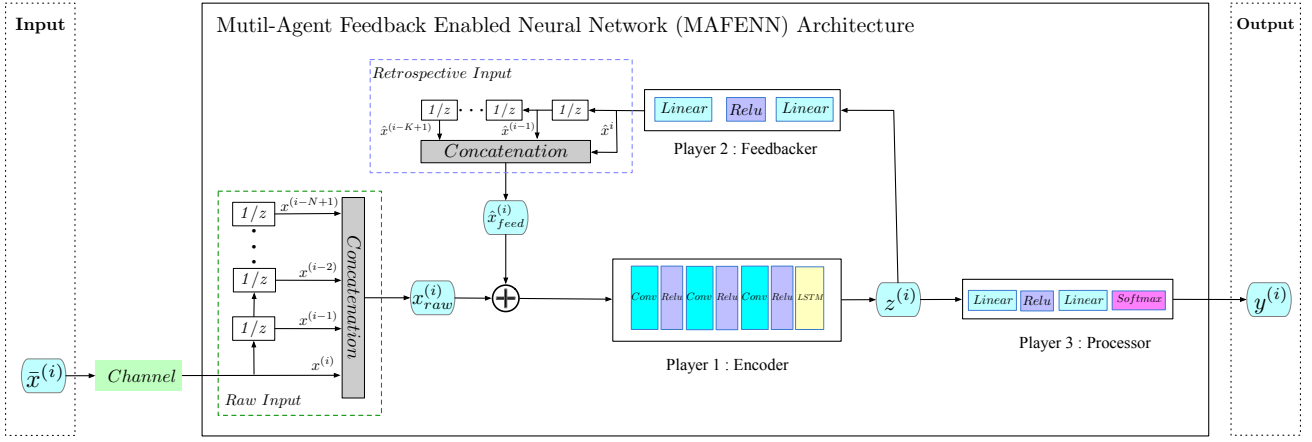}
    \caption{Model architecture of MAFENN-E with the three-player Feedback Stackelberg game.}
    \label{fig:model}
    \vspace{-4mm}
\end{figure*}

Following our preliminary exploration of multi-agent feedback networks in wireless image/video transmissions in \cite{liyangICC} and channel equalization in \cite{liyangglobalcom}, in this paper, we aim to propose the MAFENN based equalizer by introducing a feedback agent to improve feedback learning ability on feedback information with the theoretical support of the Feedback Stackelberg game.
Fig.~\ref{fig:model} illustrates the model architecture. Channel equalization can be regarded as a classification problem, where the equalizer is constructed as a decision-making device with the motivation to classify the transmitted signals as accurately as possible. 
We assume $S$ is the symbol alphabet for transmitting, and the size of $S$ is denoted as $M$. The probability that the transmitted symbol $\bar x^{(i)}=m$ from the symbol alphabet $S$ is to be determined from a training sequence, given the finite past of the received signals
\begin{equation}
    x^{(i)}_{raw}=[x^{(i)}, x^{(i-1)}, \dots, x^{(i-N+1)}]^T,
\end{equation}
where $N$ is the window length of the input signals, and the finite past of the feedback recovered signals
\begin{equation}
    \hat x^{(i)}_{feed}=[\hat x^{(i)}, \hat x^{(i-1)}, \dots, \hat x^{(i-K)}]^T,
\end{equation}
where $K$ is the window length of the feedback learning signals. 
At the time slot $i$, $\hat x^{(i)}_{feed}$ includes the recovered signals of the past $K$ time slots and one feedback learning signal in the current time slot $i$. 
Each $\hat x$ is reconstructed by Feedbacker based on the output of Encoder, which is called feedback learning process. 
So the input of MAFENN equalizer is $\dot x^{(i)}=op(x^{(i)}_{raw},\hat x^{(i)}_{feed})$. 
Then the whole MAFENN equalizer network tries to make the classification decision by the maximum conditional probability and parametrizes the conditional probability distribution as a function  in the following manner:
\begin{equation}
    f_{\theta}:\mathbb{R}^{N+K+1}\rightarrow \mathbb{R}^m \in [1,2,\dots,M],
\end{equation}
which is defined as 
\begin{equation}
    f_{\theta}(\dot x^{(i)})=P(\bar X|\dot X,{\theta})
\end{equation}
where $\theta$ represents the parameter of the whole model shown in Fig.~\ref{fig:model} and controls the probability distribution. 

Specifically, as shown in Fig.~\ref{fig:model},
Encoder maps the input $\dot x^{(i)}$ to a latent space to obtain a hidden variable $z^{(i)}$.
Different from the feedforward network that directly sends $z^{(i)}$ to the downstream network to obtain the result, we send the hidden variable $z^{(i)}$ into a designed Feedbacker. 
Feedbacker tries to reconstruct the original clean signal $\bar{x}^{(i)}$ from the hidden variable $z^{(i)}$, which aims to remove the preliminary interference. 
After a certain number of feedback learning cycles, we feed the hidden variable $z^{(i)}$ to the downstream neural network, i.e., Processor, to output the predicted class $y^{(i)}$ of the input $\bar{x}^{(i)}$. 
All the tightly cooperated three components form a multi-agent DL system to eliminate ISI.

\label{sec:system_model}
For the problem solving with our proposed Feedback Stackelberg model, we use the cross-entropy loss function and mean squared error (MSE) loss function to measure error of prediction by feedback conditional probability learning. The cross-entropy loss function is given by
\begin{equation}
l_{3}=-\frac{1}{n}\sum_{i=1}^{n} \sum_{m=1}^{M} \bar x^{(i)}_m \log y^{i}_m,
\end{equation}
where $M$ is the size of $S$, $n$ is the number of data set, $\bar x^{(i)}$ also means the ground truth class of itself, $y^{i} = f_\theta(\dot x^{(i)})$ is the output of whole framework, and $y^{i}_m$ is the probability that the i-th data belongs to the $m$-th category. 
In fact, $\bar x,\ \hat x$ are complex numbers in the system, which are equivalent to a category in the QSPK modulation mode. But for simplicity, we also denote $\bar x,\ \hat x$ as the corresponding category in cross-entropy loss function. 
The MSE loss function is given by:
\begin{equation}
\label{l2}
   l_{2} =\frac{1}{n}\left[\sum_{i=1}^{n}(\hat x^{(i)} - \bar x^{(i)})^2\right],
\end{equation}
where $\hat x$ is the output of Feedbacker.
$l_1$ is designed as the combination of $f_2$ and $f_3$, where $\beta$ is a discount constant. 
\begin{equation}
l_1=l_2+\beta l_3.
\end{equation}

Like other Stackelberg games \cite{pmlr-v119-fiez20a}, the leader begins the game by announcing its decision. Each follower executes its policies after with the full knowledge of its superior players.
Then we can assume that the follower chooses the best response to the leader's action. Then the leader is aware of this, and can utilize this information when updating its parameters. So the leader aims to solve the optimization problem as below:
\begin{equation}
\label{optimal}
\begin{aligned}
    \min_{x_1\in \mathcal{X}_1}\quad & l_1(x_1, x_2,x_3),\\
    \mathrm{s.t.}\quad & x_2 \in \arg \min_{x_2^{'} \in \mathcal{X}_2} l_2(x_1, x_2^{'}, x_3),\\
    & \mathrm{s.t.}\quad  x_3 \in \arg \min_{x_3^{'} \in \mathcal{X}_3}l_3(x_1, x_2, x_3^{'}).
\end{aligned}
\end{equation}
We noticed that $l_2$ in the above formula contains $x_3$, which is not involved in the definition of $l_2$. This is because we considered that Player 3 has an implicit effect on Player 2 during optimization. As shown in Eq.~(\ref{optimal}) and Fig.~\ref{fig:model}, Player 3 (Processor) just need to optimize itself according to other leaders' actions as a follower to others. Furthermore, Player 2 (Feedbacker) optimizes itself based on the best response of Player 3 and action of Player 1. As the top leader, Player 1 (Encoder) could update itself under the best response of Player 2 and Player 3.

\subsection{MAFENN Equalizer Components}
\label{sec:MAFENN equalizer components}
Based on the above formulation, MAFENN-E is proposed, and the detail network is shown in Fig.~\ref{fig:model}. The Encoder agent is a 4-layer pipelined neural network, in which the first part is a CNN and the subsequent part is a RNN. The original inputs are windowed raw received symbol sequences in which every symbol is constructed by the In-Phase and Quadrature (IQ) parts. We assume the window length to be $N$ thus the input for the forward part is a $N\times 2$ matrix which is suitable for 2-dimensional convolution operation. The Feedbacker agent will output a $(K+1)\times2$ matrix, where $K$ represents the feedback signals of the past $K$ slots, and  1 represents the temporary feedback value of the current slot. After concatenating the original input and feedback output, we get a $(N+K+1)\times2$ input matrix.  After the entire retrospect process is over, the signal window slides one step forwards.

The input is firstly sent to the Encoder agent composed of a CNN part and a Long Short-Term Memory (LSTM) part to extract features that are beneficial to equalization. In the CNN part, three convolutional layers and Relu activate layers are used to learn matched filters. The convolutional layers consist of a rectangular grid of neurons, where each neuron takes inputs from a rectangular section of the previous layer. The weights for this rectangular grid are the same for each neuron in the convolutional layer, moreover, they specify the convolutional filter. After each convolutional layer, a Relu activate layer is concatenated to allow complex relationships in the data to be learned. Then, we feed the learned features into a subsequent layer for temporal modeling by utilizing LSTM cells, which specifies the RNN part. Basically, the LSTM layer learns temporal dependency by memorizing the previous internal state and adding it to the current state at every single time step, and this is what the recurrent means. The weight parameters of the LSTM layer are shared across the time steps. Finally, the Encoder agent outputs latent variable $z$, which is a 128-dimensional vector.

The intuition of applying CNN and RNN as the main part of the Encoder agent is that CNN and RNN have different properties and different functions in many recovery processes in radio communication systems. We use the CNN to leverage the shift-invariant properties to learn the variance of mixing, rotation, time shifting and learn matched filters to reduce temporal variations. RNN is effective in dealing with time-dependent data and can capture the sequential information presented in the input data. Moreover, RNN can ideally implement the inverse of a finite memory system, with the result that can substantially model a nonlinear infinite memory filter. 

Subsequently, the output of the Encoder agent will be fed to the Feedbacker agent, which consists of two linear layers and a Relu activate layer. The Feedbacker agent aims to recover the original signal from the latent variable $z$ and output a 2-dimensional variable. The output of the Feedbacker agent at the current slot and the output of the past $K$ slots are concatenated with the original input and fed to a new round of feedback learning. After the feedback learning finished, the final latent variable is sent to the Processor agent, which consists of two parts, one Relu layer and a Softmax activation function to derive the outputs for the last layer. In this sense, the MAFENN framework based nerual network is trained to solve a M-class decision problem given tremendous $(x^{(i)},\hat x^{(i)})$ instances known as the training dataset.

\section{simulation results}
\label{sec:simulation}
In this section, we train the network of MAFENN-E equalizer with eleven million randomly generated digital signals. Those signals are modulated with the QPSK modulation, and we randomly select ten million as training dataset and one million as validation dataset. Our training environment is a DELL graphical work station running Ubuntu 18.04 with NVIDIA GeForce GTX 2080Ti graphical card drived by CUDA 10.0. 

Before we conduct two sets of experiments to demonstrate the performance of our proposed network, three experiments are designed firstly to optimize the framework hyperparameters,  e.g., the length of the feedback signal, the number of feedback cycles and different splicing forms of the original input and output of the Feedbacker agent. Then, we compare the SER performance among our proposed MAFENN-E based equalizer and other techniques. In the first set of experiments, the network is trained in linear multipath channels. In the second set of experiments, we test the framework in nonlinear multipath channels. 
It is worth noting that the learning dynamics involved in the convergence proof in Section~\ref{sec:convergence analysis} is not explicitly applied in our experiments due to the high time and computational complexity via such as Quasi Newton methods as in \cite{DE}.
We follow the optimization sequence described in Eq.~(\ref{optimal}), which emulates the update process of natural Stackelberg game, so we have reason to believe that the learning dynamics mentioned are implicitly applied in the training process.
In addition, our code is available in GitHub 
\footnote{https://github.com/liyang619/MAFENN\_TRANS\_2022}.

\subsection{Framework Hyperparameter Optimization}
\label{Sec:Hyperparameter Optimization}
Recalling the network of MAFENN-E consists of three agents as shown in Fig.~\ref{fig:model}, each framework hyperparameter of such a complex multi-agent system is important for the performance of the entire network. We use grid search to perform framework hyperparameter optimization, which is simply an exhaustive searching through a manually specified subset of the hyperparameter space. Before that, some of the framework hyperparameters should be manually determined to avoid a timeless searching, for example, the length of the original signal sequences $N$ is set as 12 according to the experiment result of the prior work \cite{YangLiAPCC}. In addition, the splicing forms of the original input and the Feedbacker output, the number of cycles of feedback learning and the window length of the Feedbacker output are important factors that determine the performance of the MAFENN-E equalization network. All those optimization experiments are designed in nonlinear multipath channel, which will be introduced in the next subsection in details.

\begin{figure}[htbp]
    \centering
    \includegraphics[width=0.45\textwidth]{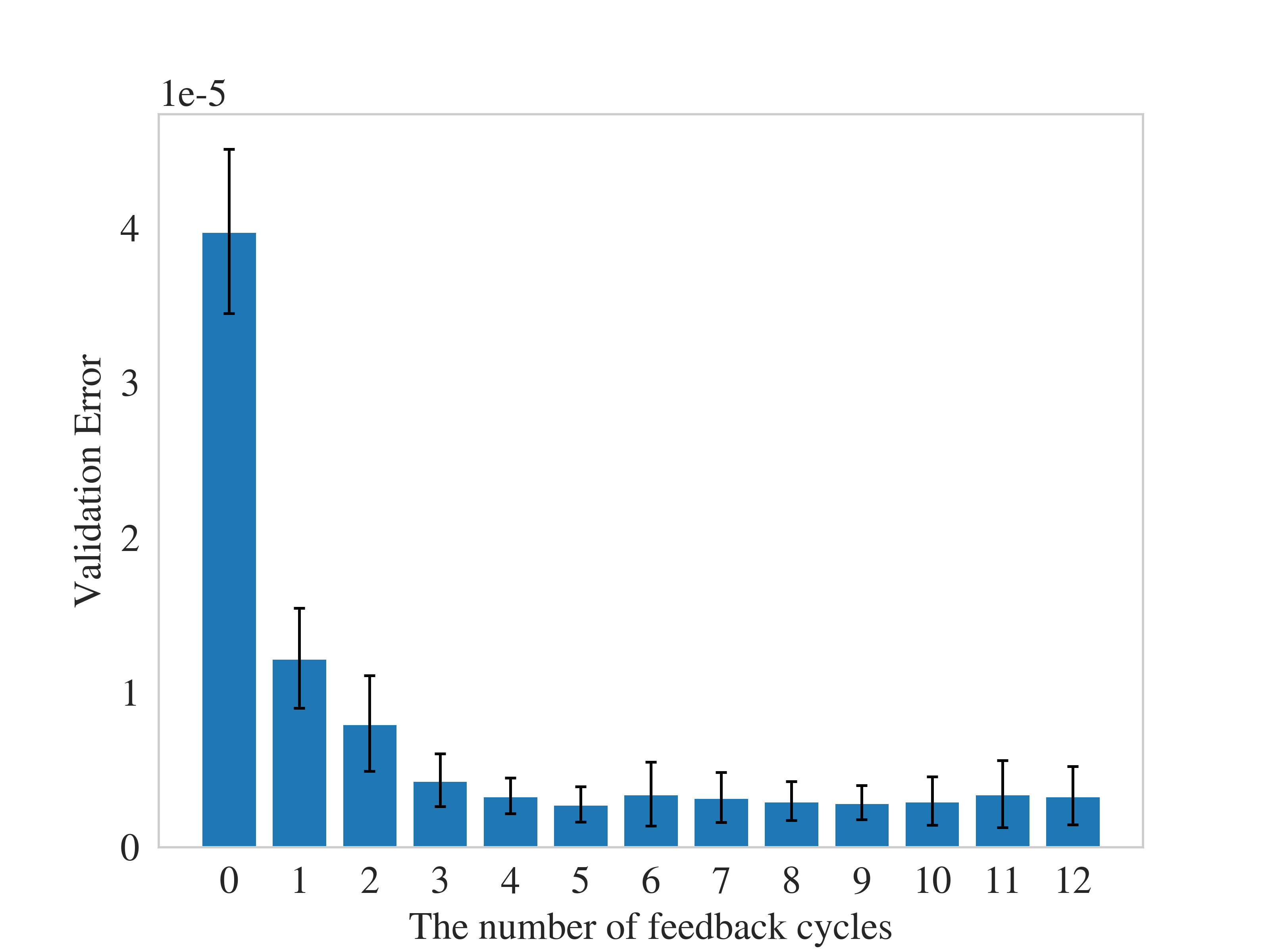}
    \caption{Mean validation error vs. the number of feedback cycles in MAFENN-E when SNR=28dB. }
    \label{fig:com_feedback_time}
\end{figure}

We use SER to reflect the validation error since it is a significant metric of channel equalizers and each of our experiments was repeated 10 times. Fig.~\ref{fig:com_feedback_time} shows the mean and standard deviation of the results when $SNR=30~dB$. Standard errors are represented in the figure by the error bars attached to each column. The chart indicates the number of feedback cycles has lower bounds for the error rate about $2\times10^{-6}$.  
In Fig.~\ref{fig:com_feedback_time}, the network becomes a  feedforward network when the number of feedback cycles is zero. While for MAFENN-E, the error rate with only one feedback cycle is also three orders of magnitude lower than that without feedback learning. With the rise of the number of feedback cycles, the error rate is close to an order of magnitude decline again. When the number of feedback cycles goes to 5, the error rate does not decrease but really fluctuates, indicating that the feedback learning has converged under the current set. So in the above experiments, considering the balance of time complexity and experimental performance, we set the number of feedback cycles to 5.

\begin{figure}[htbp]
    \centering
    \includegraphics[width=0.45\textwidth]{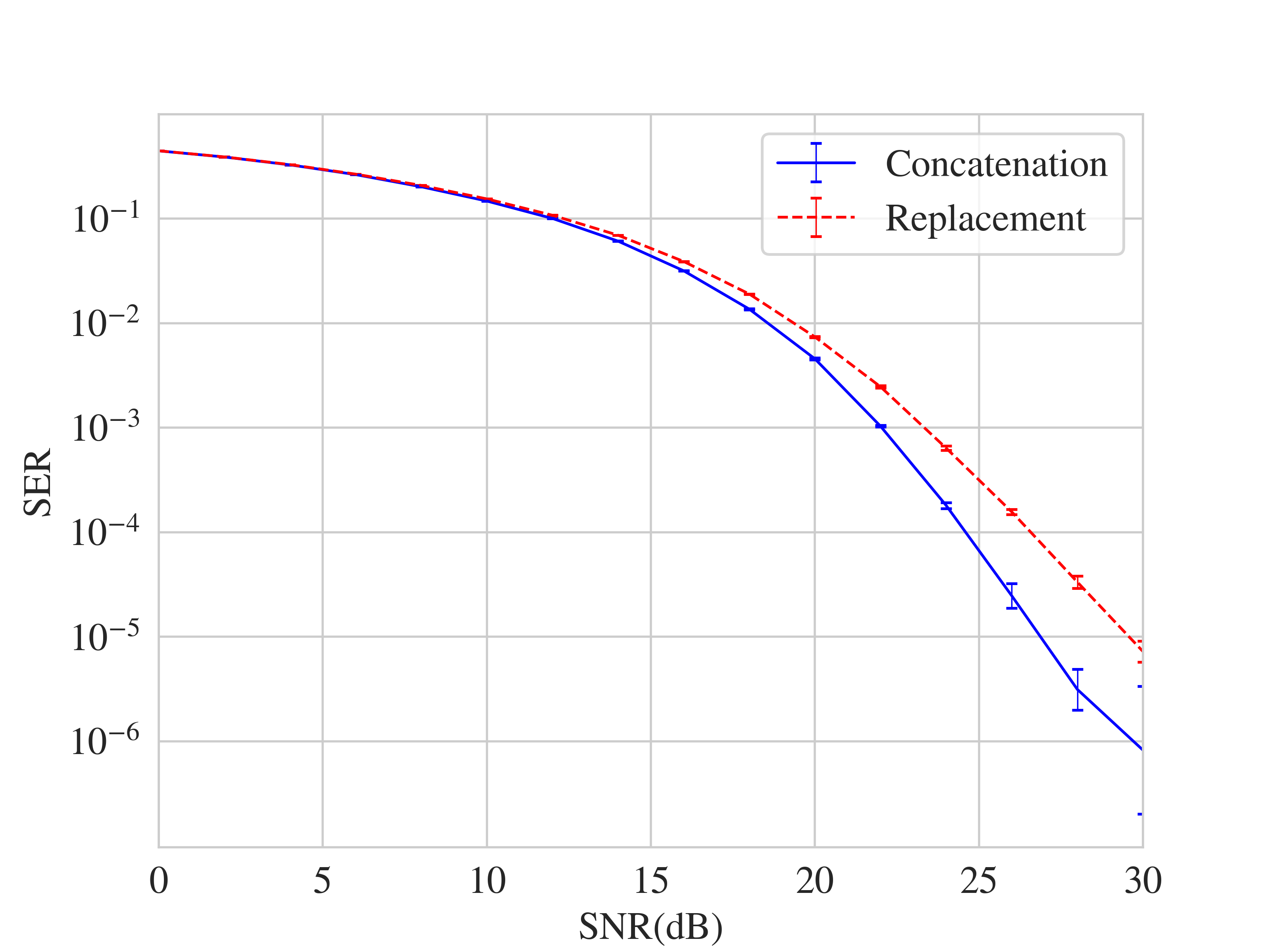}
    \caption{Mean SER comparison of different input combination forms.}
    \label{fig:com_feedback_form}
\end{figure}

In the proposed MAFENN framework, the output of the Feedbacker agent and the original input of the Encoder agent need to be combined and sent to the Encoder agent. The combination forms are crucial for the Encoder agent to extract the feature. We proposed two different combination forms. One is replacement, and the other is concatenation. The replacement method means that the original input data of the Encoder agent will be replaced by the output data of the Feedbacker agent. The concatenation method means that the original input data of the Encoder agent and the output data of the Feedbacker agent are concatenated together as the new input data of the Encoder agent, as shown in Fig.~\ref{fig:com_feedback_form}. We evaluated the SER performance of different combination methods in the SNR range from 0 dB to 30 dB. Each of our experiments was repeated 10 times, and the mean and standard deviation were shown in Fig.~\ref{fig:com_feedback_form} in the form of error bar. The solid blue line represents SER performance of the concatenation method and the dotted red line represents that of the replacement method. When the SNR value is greater than 10 dB, the SER perforamnce of the concatenation method starts to be significantly better than that of the replacement method. The SER performance of the concatenation method gains more than 3 dB when $SER=10^{-5}$. So the concatenation method is more optimal in the channel equalization problem.

\begin{figure}[htbp]
    \centering
    \vspace{-2mm}
    \includegraphics[width=0.45\textwidth]{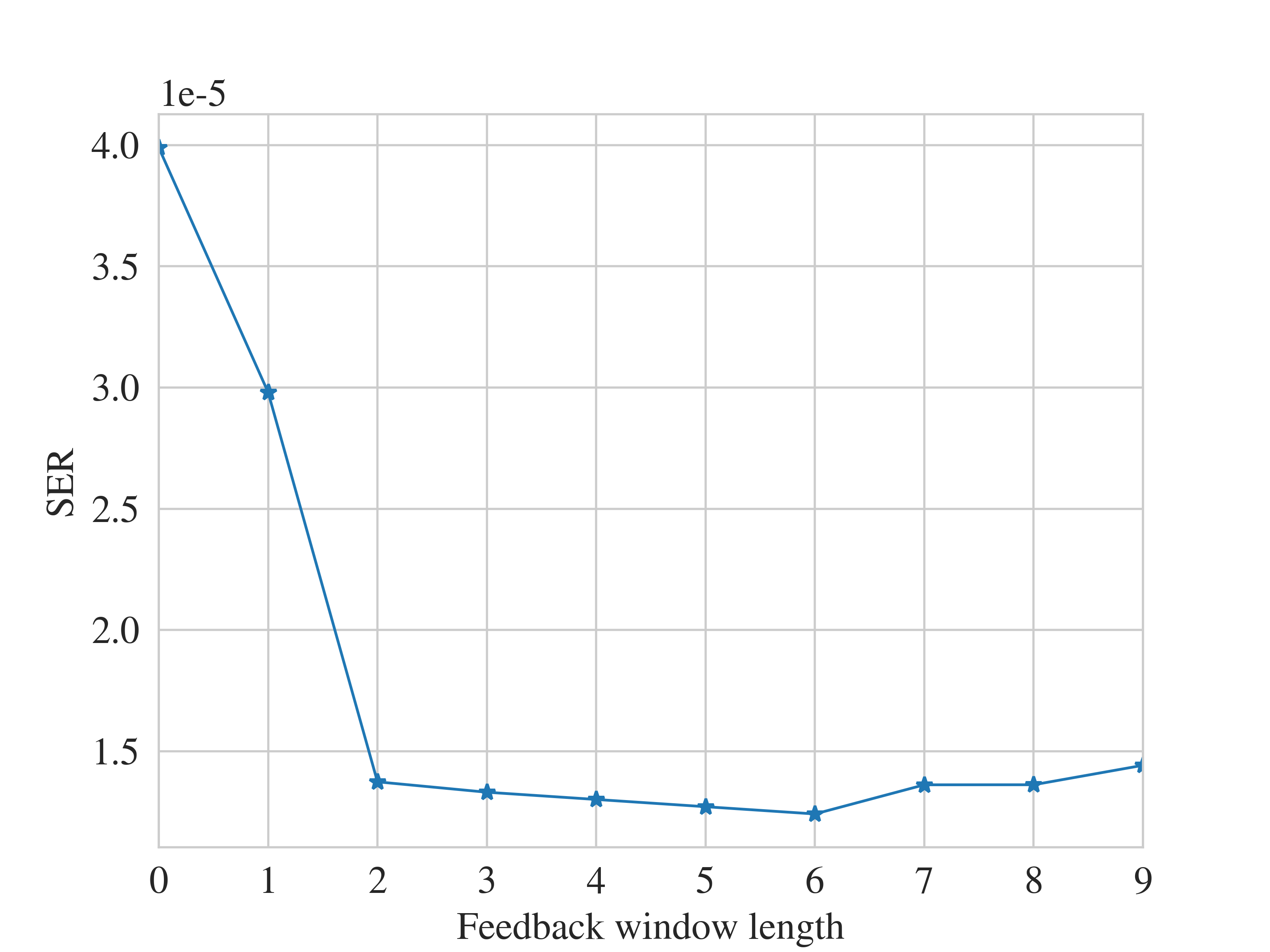}
    \caption{Mean SER vs. the window length of the feedback signal sequence in  MAFENN-E when SNR=28dB}
    \label{fig:com_retro_length}
    \vspace{-2mm}
\end{figure}

The selection of the length of the Feedbacker agent output window $K$ is also critical to the performance of the network. We selected 10 different window lengths from 0 to 9, to test which length is optimal. The result is shown in Fig.~\ref{fig:com_retro_length}, and each value represents the mean of our 10 times repeated experimental value. The value 0 represents that the network just have the current Feedbacker output without any passed feedback information. When the length is 6, the SER value curve reaches the lowest point. When the length is greater than 6, the SER performance no longer decreases but gradually increases. So $K=6$ is the optimal window length of the Feedbacker agent. 

\subsection{Experiment results in linear multipath channels}
In the first set of experiments, we evaluate the performance in linear channels with ISI and additive white Gaussian noise (AWGN). The impulse response is given in z-transform notation by:
\begin{equation}
    \label{channel model}
\resizebox{0.95\hsize}{!}{
$\begin{aligned}
H(z)&=(0.0410+j 0.0109)+(0.0495+j 0.0123) \cdot z^{-1}\\
&+(0.0672+j 0.0170)\cdot z^{-2} +(0.0919+j 0.0235)\cdot z^{-3}\\
&+(0.7920+j 0.1281)\cdot z^{-4} +(0.3960+j 0.0871)\cdot z^{-5}\\
&+(0.2715+j 0.0498) \cdot z^{-6}+(0.2291+j 0.0414)\cdot z^{-7}\\
& +(0.1287+j 0.0154)\cdot z^{-8}+(0.1032+j 0.0119) \cdot z^{-9}.
\end{aligned}
$
}
\end{equation}

\begin{figure}[htbp]
    \centering
    \includegraphics[width = 0.48\textwidth]{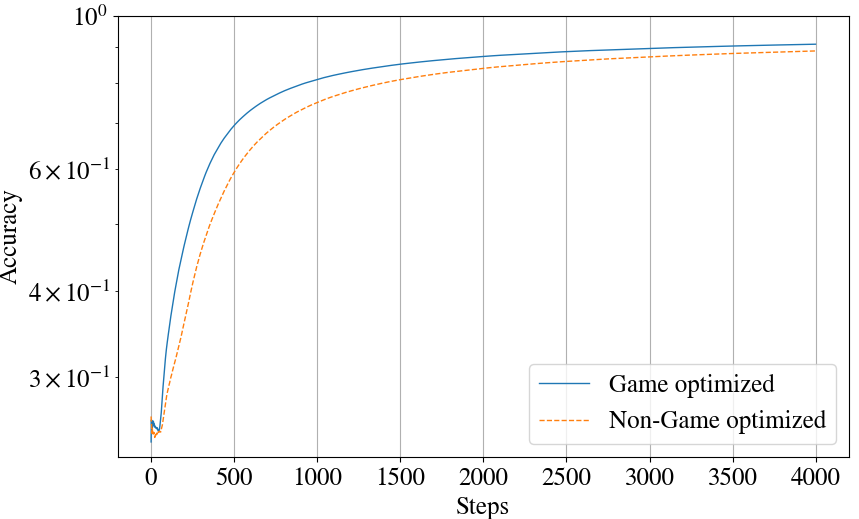}
    \caption{Comparison of the convergence speed with and  without the Stackelberg game optimization in training}
    \label{fig:conv_rate}
\end{figure}

This channel was proposed in \cite{V-channel}. We evaluate the SER performance of our MAFENN-E, RLS equalizer, MLP equalizer \cite{MLP-eq}, CRNN equalizer \cite{YangLiAPCC} and two feedback mechanism assisted DL based equalizers, i.e., MAFENN-MLP equalizer and FB-CRNN equalizer. Besides, we also conduct a two-player Stackelberg game to model a forward equalizer network based on~\cite{pmlr-v119-fiez20a} for comparison, which is named as Implicit Learning Dynamics in Stackelberg Game (ILDSG). In the MLP-MEFENN equalizer, a feedback learning agent is integrated into the MLP based equalizer networks. the FB-CRNN equalizer just adds a simple feedback path from the output to the input. All those equalizers are trained with the same training dataset, and same hyperparameters. Considering the higher computational  complexity of nonlinear equalizers as analyzed in Section \ref{sec:MAFENN equalizer challenges}, more linear equalization technologies are used in practical wireless communication systems, including ZF, MMSE, LMS and RLS, etc. In order to further reduce the computational complexity of ZF and MMSE, consider using iterative algorithms, such as LMS and RLS. RLS is prior to LMS on fast convergence speed, high estimation accuracy and good stability, and has strong adaptability to non-stationary signals. The experimental results in \cite{Ecomparison1} \cite{Ecomparison2} show that the performance of RLS is close to MMSE, while the convergence speed is higher. Therefore, this paper compares the RLS algorithm in the traditional schemes for clarification, and focuses more on comparing the performance of a variety of artificial intelligence equalization algorithms. First of all, we compare the convergence speed with and without the Stackelberg game optimization. As shown in Fig.~\ref{fig:conv_rate}, the solid blue line represents the convergence curve of the symbol decoding accuracy in the first 4000 steps with the Stackelberg game optimization, and the orange dashed line is the counterpart non-game optimized convergence curve. The simulation results show that the game-optimized MAFENN system converges faster and has a higher accuracy rate.

\begin{figure} [htbp]
    \subfigure[ Mean SER in linear channels]{
    \includegraphics[width=0.48\textwidth]{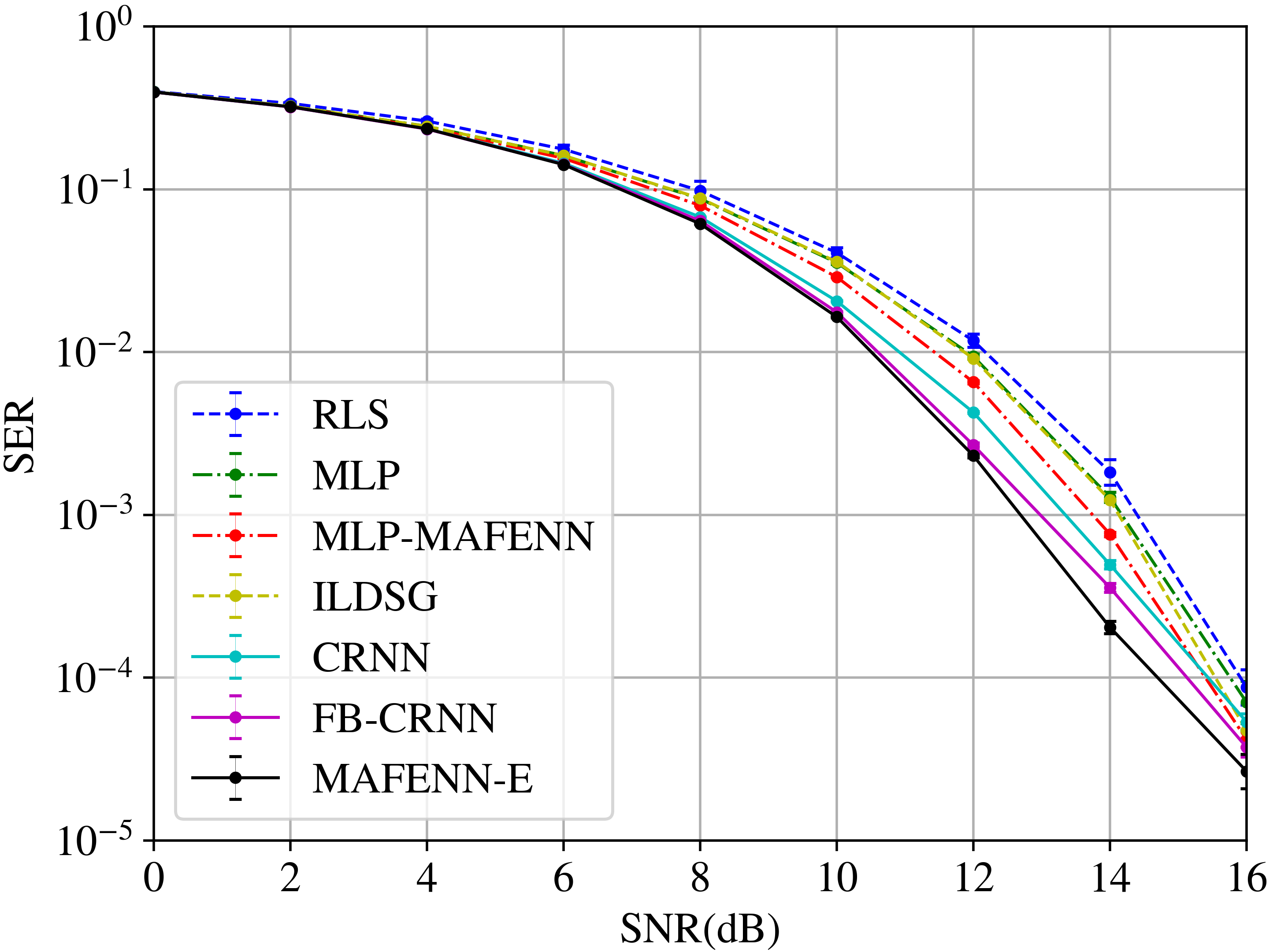}
    \label{fig:QPSK-A}
    }
    \subfigure[ Mean SER in nonlinear channels]{
    \includegraphics[width=0.48\textwidth]{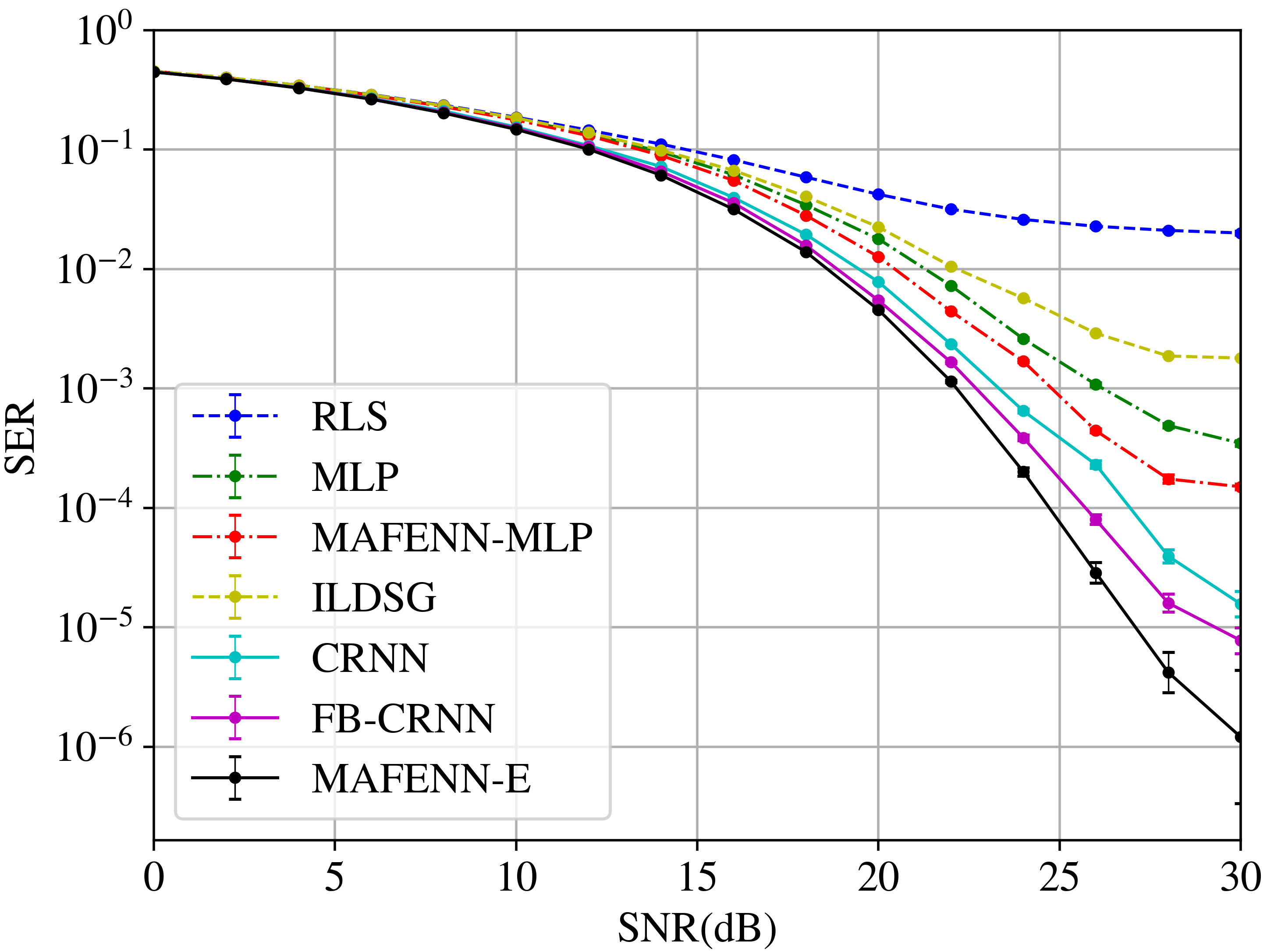}
    \label{fig:QPSK-B}
    }
    \caption{Performance comparison of the mean QPSK SER of the proposed MAFENN-E and other equalizers}
    \label{fig:QPSK}
    \end{figure}

In Fig.~\ref{fig:QPSK-A}, we provide the SER performance of those methods with regard to different SNR conditions from 0 dB to 30 dB. Each of our experiments was repeated 10 times, and the mean and standard deviation were shown in Fig.~\ref{fig:QPSK-A}. After the SNR value is greater than 8 dB, the SER performance of our proposed method has obvious gains compared to that of the other methods. Our method has about 2 dB gains over RLS when  $SER=10^{-3}$. When $SER=10^{-4}$, our method gains more than 0.5 dB compared to FB-CRNN, which just uses a feedback loop in the CRNN method. It’s worth noting that when SNR goes from 14 dB to 16 dB, the performance of MLP-MAFENN is better than that of the CRNN method. The performance of ILDSG is just a little better than that of the MLP equalizer when SNR is less than 14dB.
When SNR goes from 14 dB to 16 dB, the performance of the ILDSG equalizer gradually overtakes that of the CRNN equalizer but is still worse than that of MLP-MAFENN.
These results show that our framework is a model with better learning and generalization ability, which can achieve good results in different networks.  
Besides, 

\subsection{Experiment results in nonlinear multipath channel}
In the second sets of experiments, we perform the evaluation in nonlinear multipath channels. The nonlinearities in the communication system are mainly caused by amplifiers and mixers, which can be modeled by a nonlinear function $g(\cdot)$. The associated impulse response is the same with Eq.~(\ref{channel model}) and  the nonlinear distortion function, as used in \cite{nonliear-distortion}, is given as follows:
\begin{equation}
    |g(v)|=|v|+0.2|v|^2-0.1|v|^3+0.5 \cos(\pi|v|).
    \label{channel_distorsion}
\end{equation}
where $v$ is the signal after being transmitted through the linear channel. 

The SNR performance is plotted in Fig.~\ref{fig:QPSK-B}. Each of our experiments was repeated 10 times, and the mean and standard deviation were shown in Fig.~\ref{fig:QPSK-B}. Similar to the experiments in linear channels, we compare the SER performance of our MAFENN, RLS equalizer, MLP equalizer, ILDSG equalizer, CRNN equalizer, MAFENN-MLP equalizer, and FB-CRNN equalizer. When $SER=10^{-5}$, the performance of our method obtains more than 5 dB gains compared to the FB-CRNN method. When $SNR=30~dB$, the SER performance of our proposed equalizer is nearly an order of magnitude compared to the FB-CRNN method and more than four orders of magnitude compared  to the RLS equalizer, which shows the MAFENN framework has better performance than the traditional feedback mechanism. The performance of ILDSG equalizer is the worst network-based method in nonlinear channels. After adding our MAFENN feedback framework in the MLP method, the MAFENN-MLP also gains more than 1 dB compared to MLP equalizer. Compared to the linear channel case, our experimental results have a greater improvement, which also shows that our framework has stronger ability to face more complex channel environment.

\section{conclusion}
\label{sec:conclusion}
In this paper, we propose a multi-agent game theory based feedback structure, i.e., MAFENN. With the three fully cooperative agents, the MAFENN framework have stronger feedback learning ability and more intelligent for denoising and key information abstraction. We further formulate our MAFENN framework to a three-player Feedback Stackelberg game, and prove that the framework can converge to the Stackelberg equilibrium. Then we propose an analytical formulation of channel equalization as a conditional probability distribution learning and feedback learning problems, which can be solved by our MAFENN framework. Based on the formulation, we further propose MAFENN-E to recover and decode the transmitted signaling in wireless multipath fading channels with linear and nonlinear distortions. We compared the performance of our proposal with that of the RLS, MLP, ILDSG, MAFENN-MLP, and FB-CRNN equalizers. The experimental results show the SER performance of our proposed network outperforms that of the other methods, and is robust in either linear and nonlinear channels. As feedback mechanisms have been widely used in wireless communications to improve the estimation accuracy and scheduling performance. In the future, we will further explore our MAFENN framework to solve more problems in wireless communications such as radio resource allocation, coding/decoding to take advantage of intelligent feedback networks. Moreover, the further research may focus on integrating feedback agents into more neural networks with complicated structures, and try to build different mathematical models and solve the convergence problem.


%


\ifCLASSOPTIONcaptionsoff
  \newpage
\fi



\bibliographystyle{IEEEtran}
%


\bibliography{reference}

\end{document}